\documentclass[sigchi]{acmart}

\setcopyright{none}
\renewcommand\footnotetextcopyrightpermission[1]{} 

\usepackage{balance}

\begin{document}

\title{Design-Space Exploration of SNN Models using Application-Specific Multi-Core Architectures}




\author{Sanaullah}
\authornotemark[1]
\affiliation{%
  \institution{ Dep. of Engineering and Mathematics, Bielefeld University of Applied Science, Bielefeld, Germany}
}

\author{Shamini Koravuna}
\affiliation{%
  \institution{Center for Cognitive Interaction Technology, University of Bielefeld,
Bielefeld, Germany}
}

\author{Ulrich Rückert}
\affiliation{%
  \institution{Center for Cognitive Interaction Technology, University of Bielefeld,
Bielefeld, Germany}
}

\author{Thorsten Jungeblut}
\affiliation{%
  \institution{ Dep. of Engineering and Mathematics, Bielefeld University of Applied Science, Bielefeld, Germany}
}

\maketitle

\section{Introduction}


In recent years, there have been a growing number of experimental results documenting the importance of spike timing for neural computations, which have driven the development of spiking neural networks (SNNs). This third generation of (spiking) neural networks is considered energy-efficient and expensive in terms of computation compared to artificial neural networks (ANNs). SNNs provide many benefits, but they are also challenging to implement due to complex neural computational structures \cite{20}. Although there has been significant development in dedicated simulators for analyzing and visualizing SNN behavior. The most cutting-edge SNN simulators, for example, Brian2 \cite{2}, NEST \cite{3}, and CARLsim \cite{4}, were designed primarily for studying brain functions and neuronal dynamics, but due to the use of certain simulators, these tools are not user-friendly, additional functions must be specified in low-level programming languages like C++ and integrated into the simulator code. Sometimes, simulators require users to program in a domain-specific language, for example, the NEST simulator requires NESTML, and NMDOL is required for the NEURON simulator \cite{5}. Such simulators allow the users to acquire simulations precisely in a relatively short period as compared to ANNs \cite{24,25}. Nevertheless, there are many challenges and computational issues related to SNN. For example, the model must initialize with the accurate biological representations of the neurons. However, the present simulators require a lot of time and extensive amounts of code for constructing the neural network designs as well as for evaluating and visualizing their behavior, hence an interactive simulator with little to no code is needed given all these limitations. Furthermore, a fast, run-time interaction, visualization, and analysis-based simulator not only helps to accelerate the simulation process but will also speed up the designing, prototyping, and parameter tuning \cite{23}. Additionally, the whole area also benefits from studying other algorithms, such as training advances in the whole field. With this motivation and the difficulties that currently exist in comprehending and utilizing the promising features of SNNs, we proposed a novel run-time multi-core architecture-based simulator called "RAVSim" (Runtime Analysis and Visualization Simulator) \cite{6}, a cutting-edge SNN simulator, developed using LabVIEW \cite{7} and it is publicly available on their website as an official module \cite{8}. RAVSim is a runtime virtual simulation environment tool that enables the user to interact with the model, observe its behavior of output concentration, and modify the set of parametric values at any time while the simulation is in execution. Recently some popular tools have been presented, but we believe that none allow users to interact with the model simulation in run time.



\section{Project Description}

Our project aims to analyze resource-efficient implementations of biologically inspired spiking neural networks, which on the one hand, enable the execution of SNNs in a resource-efficient manner and on the other hand, enable the possibility of online learning adaptation. The primary focus of the project is to explore the design space for potential computer vision applications (i.e., object detection/recognition). Artificial intelligence activities will employ a variety of applications, including one that uses an online learning strategy for object detection and recognition. In the first phase, we'll study hardware architectures and a resource-efficient simulator that lets users run the model analyzing and visualization of the behavior using any reconfigurable computing platform such as CPU/GPU. Furthermore, our primary goal is to maximize performance by utilizing a CPU-based hardware platform using a multi-core architecture with configurable processors. Second, several neurons and synapse models will be assessed, along with their configurations. To perform in-silico analysis and simulation, it is extremely important to understand the different parameters of input-output concentration to obtain precise and accurate parametric values, i.e., what values are required to obtain a stable output concentration and which parametric values are dependent on one another in each event. Even with the moderate size of the set of neurons, defining and simulating the model using the combinations of these input concentrations may need highly precious balanced values. Alternatively, a run-time simulator offers a more useful method for simulating SNNs, allowing users to directly alter the simulation at any time to see how the model responds to changes in output concentration. This type of simulator not only provides an environment where users can completely comprehend this complicated structural model in an approachable way, but it also enables the SNN community to evaluate and visualize the model easily by raising or lowering the input at any level or instant in time. Also, it makes possible different learning processes that would not otherwise be easily possible.

To this end, we present a resource-efficient runtime simulator known as the RAVSim. Spiking neural model analysis and visualization in RAVSim helps users to extract a balanced set of parametric values and allows users to interact with the model in a real-time simulation while viewing the SNN parameter reactions graphically. This process is analogous to setting up parameters for experimentation and testing a model in any other programming language like Python. But to fully understand these parameters' output behavior, we need to write a lengthy code, fine-tune the model with different parameter values, and repeatedly run the entire program several times with a large number of executable files to observe how they behave and find the proper stable values of these hyperparameters. On the other hand, RAVSim is a replacement for these time-consuming code-based experiments in both prospective studying and comprehension, where the model may need to run with several set parametric values, but only a handful of them will work for the following reasons:

\begin{itemize}
    \item Optimal parametric value and their balancing are necessary for the SNNs model.

    \item A model is constructed by combining various components, and the function of each component depends on various factors like a threshold value, membrane capacitance, and membrane time constant.

   \item These models' multiple states—their response to various inputs or how they alter in response to various inputs—define them.

\end{itemize}

\section{Future Directions}

A complete understanding of these complex SNN model mechanisms can only be achieved through run-time interaction with the simulation. We have stated some of the main architectural elements that distinguish RAVSim as a standalone simulator. These characteristics demonstrate that RAVSim strikes a unique compromise between the competing demands for flexibility, a user-friendly user interface, and accelerated performance, allowing users to compare the outcomes to the available simulators. We tested our implemented algorithms in the presented simulator using a ”real-world” simulation-based software testing method. In addition, the experimental results indicate that the proposed SNN model analysis and simulation algorithm is significantly faster at estimating and visualizing the behavior of a neural network (SNN). The discussion of the implemented algorithm and RAVSim architecture can be found in \cite{6} and details of each experimental result can be accessed in \cite{9}. For future work, we are continuously working on improving RAVSim. Additionally, by implementing other SNN neuron and synapse models, and various learning techniques, we are also working on a real-time computer vision application implementation with event-based cameras, where users are allowed to control the application characteristics at runtime to analyze the model output. 

\subsection{Availability}
RAVSim is an open-source simulator and it is published on LabVIEW’s official website \cite{8}. The user manual guide can be downloaded at \cite{9} and a video demonstration of RAVSim can be accessed at \cite{22}

\subsection{ACKNOWLEDGEMENTS}
This research was supported by the research training group "Dataninja” (Trustworthy AI for Seamless Problem Solving: Next Generation Intelligence Joins Robust Data Analysis) funded by the German federal state of North Rhine-Westphalia and the project SAIL. The project SAIL is receiving funding from the program "Netzwerke 2021", an initiative of the Ministry of Culture and Science of the State of North Rhine-Westphalia Westphalia. The sole responsibility for the content of this publication lies with the authors.

\balance

\end{document}